\documentclass[runningheads]{llncs}

\usepackage{graphicx}
\usepackage{amsmath}
\usepackage{amssymb}
\usepackage{tikz}
\usetikzlibrary{shapes.geometric, arrows, positioning, fit, calc}
\usepackage{hyperref}
\usepackage{csquotes}
\usepackage[capitalize,nameinlink]{cleveref}
\usepackage{acronym}

\crefname{figure}{Fig.}{Figs.}
\Crefname{figure}{Fig.}{Figs.}
\crefname{equation}{Eq.}{Eqs.}
\Crefname{equation}{Eq.}{Eqs.}

\newacro{RL}{reinforcement learning}
\newacro{DRL}{deep reinforcement learning}
\newacro{CPS}{cyber-physical system}
\newacroplural{CPS}[CPSs]{cyber-physical systems}
\newacro{MPC}{model predictive control}
\newacro{MDP}{Markov decision process}
\newacro{CMDP}{constrained Markov decision process}
\newacro{CPO}{constrained policy optimization}
\newacro{CBF}{control barrier function}
\newacro{1-DOF}{1-degree of freedom}
\newacro{PPO}{proximal policy optimization}
\newacro{GP}{Gaussian process}
\newacro{MORL}{multi-objective reinforcement learning}
\newacro{DLL}{dynamic link library}

\begin{document}

\title{Safe Reinforcement Learning using Ideas from Model Predictive Control}
\titlerunning{Safe RL using Ideas from MPC}

\author{Georg Sch\"afer\inst{1,2} \and
Jakob Rehrl\inst{1} \and
Stefan Huber\inst{1} \and
Simon Hirlaender\inst{2}}

\authorrunning{G. Sch\"afer et al.}

\institute{Josef Ressel Centre for Intelligent and Secure Industrial Automation, \\
Salzburg University of Applied Sciences, Salzburg, Austria \and
Department of Artificial Intelligence and Human Interfaces, \\
Paris Lodron University of Salzburg, Salzburg, Austria\\
\email{georg.schaefer@fh-salzburg.ac.at}}

\maketitle

\begin{abstract}
\Ac{RL} enables the synthesis of control policies directly from data, making it highly appealing for complex \acp{CPS} and robotics.
A persistent challenge, however, is ensuring strict, hard safety constraints during the active learning phase.
In real-world physical systems, violating mechanical limits can cause irreversible damage, necessitating that exploration remains strictly within safe operational regions.
We propose a generalized framework that combines the adaptive, high-performance nature of \ac{DRL} with the formal safety guarantees of \ac{MPC}.
Using a mathematical model of the system dynamics, offline \ac{MPC} computations define a feasible state-action space, representing all safe combinations of system states and control inputs that guarantee constraint satisfaction.
During training and deployment, the \ac{RL} agent's instantaneous actions are projected onto this globally verified feasible set via a safety filter.
We systematically evaluate our generalized approach on a non-linear \ac{1-DOF} laboratory testbed, demonstrating successful exploration and stable policy convergence on physical hardware.
\end{abstract}
\acresetall

\section{Introduction}

The application of \ac{RL} to \acp{CPS} and industrial robots holds immense promise.
\Ac{RL} algorithms learn optimal control policies through active interaction with an environment, offering high potential to solve complex, non-linear control problems where traditional analytical models or linear approximations may fall short \cite{sutton2018reinforcement,kober2013reinforcement}.
However, standard \ac{RL} fundamentally relies on extensive, unconstrained exploration of the state-action space to discover these optimal policies. 

This core requirement creates a critical safety bottleneck: pure \ac{RL} agents optimize for scalar reward signals but do not inherently respect strict physical safety constraints~\cite{garcia2015comprehensive}.
In the context of real-world physical systems, such unconstrained exploration is unacceptable.
Violating mechanical limits, thermal bounds, or actuator constraints can lead to hardware failure, costly downtime, or irreversible system damage.
Therefore, an agent must avoid what we define as the \textit{unsafe state}: any situation where the system either directly violates constraints or passes a \enquote{point of no return} where a future constraint violation becomes physically inevitable due to system dynamics, regardless of any subsequent control actions.

Bridging the gap between the theoretical promise of \ac{RL} and safe physical deployment requires mechanisms that guarantee absolute constraint satisfaction throughout the entire learning process.
We propose a generalized framework combining the data-driven learning capabilities of \ac{RL} with the formal, forward-looking safety guarantees of offline \ac{MPC}.
Rather than evaluating safety purely online, which is often computationally prohibitive for the high-frequency control loops required in modern \acp{CPS}, our approach uses offline \ac{MPC} as an oracle to pre-define a \enquote{feasible state-action space} ($\mathcal{F}$).
During training, a deterministic projection filter maps any potentially unsafe action proposed by the \ac{RL} agent back into this safe set, avoiding violations during both exploration and deployment.

\section{Related Work}

The challenge of safe exploration in \ac{RL} has garnered significant attention, generally falling into modifications of the optimization objective (soft constraints) or architectural interventions (hard constraints).

\textbf{Constrained Markov Decision Processes:}
%A foundational approach to safe \ac{RL} involves framing the problem as a \ac{CMDP} \cite{altman2021constrained}, where the agent maximizes expected rewards subject to expected cost constraints.
A foundational approach to safe \ac{RL} involves framing the problem as a \ac{CMDP} \cite{altman2021constrained}, where the agent maximizes expected rewards while ensuring that the expected cumulative cost, representing the likelihood of unsafe behavior, remains below a predefined threshold.
Algorithms such as \ac{CPO} \cite{achiam2017constrained} and Lagrangian relaxation methods have been widely adopted to solve \acp{CMDP}.
While effective for soft constraints, these methods optimize for expected constraint satisfaction over a temporal horizon.
Consequently, they cannot provide the absolute, deterministic, step-by-step safety guarantees required to prevent catastrophic failure on physical hardware during the highly stochastic initial stages of learning \cite{garcia2015comprehensive}.

\textbf{Lyapunov-based and Reachability Approaches:}
To introduce safety constraints, researchers have explored Lyapunov-based safe \ac{RL}, constructing continuous Lyapunov functions to restrict the policy update step to a safe subset of policies \cite{chow2018lyapunov}.
Alternatively, reachability analysis computes safe backward reachable sets.
Fisac et al. \cite{fisac2018general} proposed a general safety framework using Hamilton-Jacobi reachability to override the learning agent when it approaches the boundary of the safe set.
While providing strong theoretical guarantees, computing exact reachable sets suffers heavily from the curse of dimensionality and is often intractable for complex continuous systems.

\textbf{Reactive Shielding and \Acp{CBF}:}
Architectural interventions often utilize a separate mathematical construct to override the \ac{RL} agent online.
Shielding \cite{alshiekh2018safe} relies on formal verification to construct an automaton that monitors the system and intervenes when a safety property is threatened.
Similarly, \acp{CBF} \cite{ames2019control,cheng2019end} synthesize a safe control set online by ensuring the derivative of a barrier function remains within acceptable bounds, allowing end-to-end safe \ac{RL}. 

%However, it is crucial to note that standard \ac{DRL} policies act \textit{instantaneously} on the current state information.
%They are not reactive in a delayed sense.
%If a safety mechanism relies purely on reacting to an already-chosen action online, it may intervene too late, particularly in systems with high inertia where the system has already entered a trajectory toward an inevitable collision.
However, it is crucial to note that standard \ac{DRL} policies lack a predictive horizon, mapping current state information directly to instantaneous actions.
If an online safety mechanism evaluates these proposed actions without simulating the long-term system dynamics, it may intervene too late, particularly in systems with high inertia where the system has already entered a trajectory toward an inevitable collision.
Our approach contrasts with online \acp{CBF} and reactive shielding by pre-computing the recursive feasibility boundaries offline using \ac{MPC}.
This ensures that the instantaneous action requested by the \ac{DRL} policy is immediately projected into a globally verified, recursively safe space before any mechanical action is initiated, shifting the heavy computational burden out of the real-time control loop.

\section{Theoretical Background}

To formalize the integration of data-driven learning and model-based safety, we first establish the theoretical foundations of both domains.

\subsection{Reinforcement Learning and Markov Decision Processes}
In the standard \ac{RL} framework, the interaction between an agent and a dynamic environment is modeled as an infinite-horizon, discounted \ac{MDP} \cite{sutton2018reinforcement}.
An \ac{MDP} is formally defined by the tuple $\mathcal{M} = (\mathcal{S}, \mathcal{A}, \mathcal{P}, r, \gamma, \mu)$, where:
\begin{itemize}
    \item $\mathcal{S} \subseteq \mathbb{R}^n$ is the continuous state space.
    \item $\mathcal{A} \subseteq \mathbb{R}^m$ is the continuous action space.
    \item $\mathcal{P}: \mathcal{S} \times \mathcal{A} \times \mathcal{S} \rightarrow [0, \infty)$ is the transition probability density function, governing the environment dynamics such that $s_{t+1} \sim \mathcal{P}(\cdot | s_t, a_t)$.
    \item $r: \mathcal{S} \times \mathcal{A} \rightarrow \mathbb{R}$ is the scalar reward function.
    \item $\gamma \in [0, 1)$ is the discount factor, prioritizing immediate rewards over delayed ones.
    \item $\mu$ is the initial state distribution.
\end{itemize}

The objective of the \ac{RL} agent is to learn a parameterized, stochastic policy $\pi_\theta(a|s)$ (or a deterministic policy $\pi_\theta(s)$) that maximizes the expected cumulative discounted return:
\begin{equation}
    J(\pi_\theta) = \mathbb{E}_{\tau \sim \pi_\theta} \left[ \sum_{t=0}^{\infty} \gamma^t r(s_t, a_t) \right],
\end{equation}
where the trajectory $\tau = (s_0, a_0, s_1, a_1, \dots)$ is generated by sampling $s_0 \sim \mu$, $a_t \sim \pi_\theta(\cdot | s_t)$, and $s_{t+1} \sim \mathcal{P}(\cdot | s_t, a_t)$ \cite{SKRHH25}. 

\subsection{Model Predictive Control}
The standard discrete-time \ac{MPC} optimization problem is formulated as minimizing a cost function:
\begin{align}
    \min_{\mathbf{U}} \quad & \sum_{i=0}^{H-1} \ell(x_{k+i|k}, u_{k+i|k}) \label{eq:mpc_cost} \\
    \text{subject to:} \quad & x_{k+i+1|k} = f(x_{k+i|k}, u_{k+i|k}), \quad \forall i \in \{0, \dots, H-1\}, \label{eq:mpc_dyn} \\
    & x_{k+i|k} \in \mathcal{X}_{\text{safe}}, \quad \forall i \in \{1, \dots, H\}, \label{eq:mpc_state_const} \\
    & u_{k+i|k} \in \mathcal{U}_{\text{safe}}, \quad \forall i \in \{0, \dots, H-1\}, \label{eq:mpc_input_const}
\end{align}
where $\ell(\cdot, \cdot)$ is the stage cost and $f(\cdot, \cdot)$ represents the discrete-time system dynamics.
\Cref{eq:mpc_state_const,eq:mpc_input_const} enforce hard state and input constraints. 

While classical \ac{MPC} theory often requires the computation of a control invariant terminal set to formally guarantee recursive feasibility and stability, determining such a set can be highly complex for non-linear systems.
In practical applications, evaluating the constraints over a sufficiently large prediction horizon $H$ serves as an effective and computationally tractable alternative to ensure system safety.

\section{Generalized Problem Formulation}

To bridge the gap between control theory and \ac{RL}, we must formally define the system dynamics, the constraints, and the learning objective.

\subsection{Continuous System Dynamics}
We consider a generic cyber-physical system whose dynamics are governed by a continuous-time non-linear differential equation:
\begin{equation}
    \dot{x}(t) = f_{\text{cont}}(x(t), u(t)),
\end{equation}
where $x(t) \in \mathcal{X} \subset \mathbb{R}^n$ represents the system state and $u(t) \in \mathcal{U} \subset \mathbb{R}^m$ represents the continuous control input vector. 

\subsection{Operational Constraints and the Discrete \Ac{MDP}}
Real-world physical systems are subject to strict operational and mechanical constraints.
Let $\mathcal{X}_{\text{safe}} \subset \mathcal{X}$ denote the set of physically allowable states (e.g., position and velocity boundaries) and $\mathcal{U}_{\text{safe}} \subset \mathcal{U}$ denote the hard physical actuator limits (e.g., maximum motor voltage or torque).
The system must operate strictly within $\mathcal{X}_{\text{safe}} \times \mathcal{U}_{\text{safe}}$.

For the \ac{RL} framework, these continuous dynamics are discretized over a finite sample time $\Delta t$, yielding a deterministic transition function:
\begin{equation}
    s_{t+1} = f_{\text{disc}}(s_t, a_t) = s_t + \int_{t}^{t+\Delta t} f_{\text{cont}}(s(\tau), a(\tau)) \, d\tau,
\end{equation}
where $s_t = x(k \Delta t)$ and $a_t = u(k \Delta t)$.
This maps the continuous control problem to the underlying discrete \ac{MDP} required for the agent.

\subsection{The Safe Exploration Paradox}
The objective of the \ac{DRL} agent is to learn a parameterized policy $\pi_\theta(s_t)$ that maximizes the expected discounted return.
However, standard policy optimization fundamentally relies on stochastic exploration mechanisms, such as adding Gaussian noise to the continuous action space, where $a_t \sim \mathcal{N}(\pi_\theta(s_t), \Sigma)$.

This creates a safety paradox: stochastic exploration will inevitably propose actions that push the system outside of $\mathcal{X}_{\text{safe}} \times \mathcal{U}_{\text{safe}}$.
Furthermore, satisfying immediate, single-step constraints (i.e., merely ensuring that the next state transition $s_{t+1} \in \mathcal{X}_{\text{safe}}$) is wholly insufficient.
Due to system inertia, the agent must avoid entering states from which all available control authority in $\mathcal{U}_{\text{safe}}$ is mathematically insufficient to prevent a future violation.
This necessitates a mechanism that guarantees recursive, forward-looking safety without halting the active exploration process.

\section{Safe Exploration via Offline \Ac{MPC}}

To resolve this paradox without impeding the real-time control loop, we define the true feasible state-action space, $\mathcal{F}$, which guarantees recursive, long-term physical safety.

Let $\mathcal{S}^+ \subseteq \mathcal{X}_{\text{safe}}$ denote the admissible state space, and $\mathcal{S}^- = \mathcal{S} \setminus \mathcal{S}^+$ denote the unsafe state space.
For any state $s \in \mathcal{S}^+$, a control action is only truly safe if there exists an infinite sequence of future actions within $\mathcal{U}_{\text{safe}}$ that keeps the system out of $\mathcal{S}^-$.
We denote this set of safe actions for a given state $s$ as $\mathcal{F}_{\mathcal{A}}(s) \subseteq \mathcal{U}_{\text{safe}}$. 

We define the feasible state space ($\mathcal{F}_{\mathcal{S}}$) as the set of all states where at least one safe action exists:
\begin{equation}
    s \in \mathcal{F}_{\mathcal{S}} \Leftrightarrow \mathcal{F}_{\mathcal{A}}(s) \neq \emptyset.
\end{equation}
The global feasible state-action space $\mathcal{F}$ is the union of all safe state-action pairs:
\begin{equation}
    \mathcal{F} = \bigcup_{s \in \mathcal{F}_{\mathcal{S}}} \{s\} \times \mathcal{F}_{\mathcal{A}}(s).
\end{equation}

\subsection{Model Predictive Control as a Feasibility Oracle}
We employ the constrained \ac{MPC} optimization formulated in \crefrange{eq:mpc_cost}{eq:mpc_input_const} to approximate infinite-horizon safety.
For a given state-action pair $(s,a)$, the \ac{MPC} oracle simulates the system dynamics forward over the finite horizon $H$, forcing the first action in the sequence to be $u_{k|k} = a$. 

To circumvent the need for a rigorously computed terminal set, we utilize a conservatively large prediction horizon $H$.
For mechanical systems, if $H$ is chosen such that the prediction window significantly exceeds the maximum time required to bring the system to a complete halt using maximum opposing control effort, surviving for $H$ steps acts as a practical guarantee for long-term safety.
Therefore, if the optimizer can find a physically valid sequence of subsequent actions $\mathbf{U}$ that respects all constraints over this sufficiently large horizon $H$, the state-action pair is deemed safe.

We define the binary oracle function mathematically as:
\begin{equation}
    \text{Oracle}(s, a) = 
    \begin{cases} 
      1, & (s, a) \in \mathcal{F}, \\
      0, & (s, a) \notin \mathcal{F}.
    \end{cases}
\end{equation}
By executing this oracle strictly offline, we can evaluate complex, non-linear trajectories over long horizons $H$ without violating the strict latency requirements of the online \ac{RL} policy.

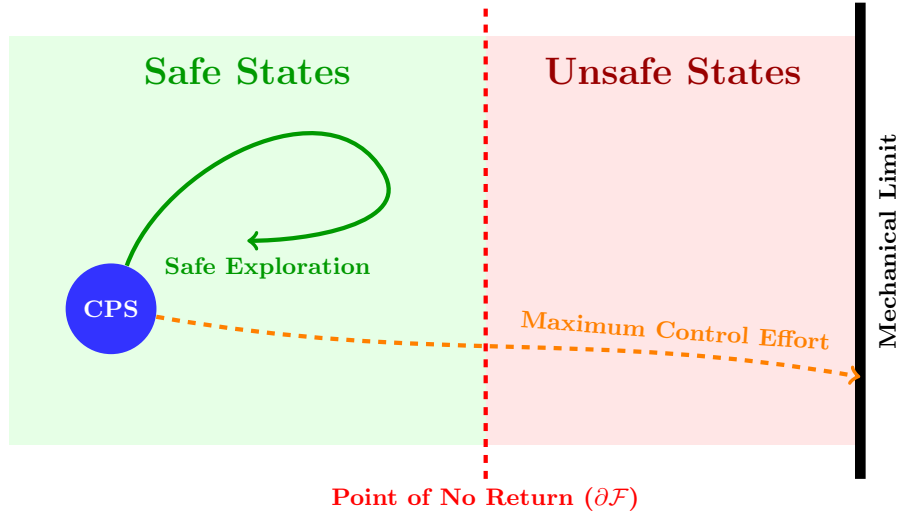
\begin{figure}[htbp]
\centering
\begin{tikzpicture}[scale=0.9, node distance=2cm]
    % Expanded Safe Region (Wider to the left)
    \fill[green!10] (-6, -3) rectangle (1, 3);
    \node[green!60!black, font=\Large\bfseries] at (-2.5, 2.5) {Safe States};
    
    % Expanded Unsafe Region (Wider to the right)
    \fill[red!10] (1, -3) rectangle (6.5, 3);
    \node[red!60!black, font=\Large\bfseries] at (3.75, 2.5) {Unsafe States};
    
    % Boundary (Point of no return)
    \draw[red, dashed, ultra thick] (1, -3.5) -- (1, 3.5);
    \node[red, font=\bfseries] at (1, -3.8) {Point of No Return ($\partial \mathcal{F}$) };
    
    % Mechanical Limit moved further right to match the new width
    \draw[black, line width=4pt] (6.5, -3.5) -- (6.5, 3.5);
    \node[black, font=\bfseries, rotate=90] at (6.9, 0) {Mechanical Limit};
    
    % CPS Node (Moved further down)
    \node[circle, fill=blue!80, text=white, font=\bfseries, minimum size=1.2cm] (cps) at (-4.5, -1.0) {CPS};
    
    % Safe Trajectory (Adjusted to stay lower and avoid the title)
    \draw[->, green!60!black, ultra thick] (cps) to[out=70, in=120] (-0.5, 1.0) to[out=-60, in=0] (-2.5, 0.0);
    \node[green!60!black, font=\bfseries] at (-2.2, -0.4) {Safe Exploration};
    
    % Unsafe Trajectory attempt (Text moved to pos=0.75 and given a white background to prevent line overlap)
    \draw[->, orange, dashed, ultra thick] (cps) to[out=-10, in=170] 
        node[pos=0.75, sloped, above=3pt, inner sep=2pt, font=\bfseries, text=orange] {Maximum Control Effort} 
        (6.5, -2.0);
\end{tikzpicture}
\caption{Abstract visualization of the state space. Actions must be restricted before crossing the ``Point of No Return'' ($\partial \mathcal{F}$). Beyond this boundary, even maximum control effort is physically insufficient to prevent a collision with hard constraints.}
\label{fig:point_of_no_return}
\end{figure}

\subsection{Scaling the Offline Mapping via System Convexity}
To construct the offline map of $\mathcal{F}$, we employ a systematic grid search across the discretized state-action space $(\mathcal{S} \times \mathcal{A})$.
For a given state space discretization, evaluating every possible continuous action $a \in \mathcal{A}$ via the expensive \ac{MPC} oracle is computationally highly inefficient. 

However, we drastically reduce the computational complexity by exploiting the physical properties of many continuous dynamical systems.
%For mechanical systems with monotonic control authority (such as voltage applied to a DC motor, or thrust applied to a rotor), the set of feasible actions $\mathcal{F}_{\mathcal{A}}(s)$ for a given state $s$ forms a continuous, convex interval. 
Specifically, for mechanical systems modeled with control-affine (or input-affine) dynamics, where the control input enters the system equations linearly (such as voltage applied to a DC motor), the safety constraints map to linear inequalities with respect to the action space \cite{ames2019control,khalil2002nonlinear}.

Consequently, the set of feasible actions $\mathcal{F}_{\mathcal{A}}(s)$ for a given state $s$ strictly forms a continuous, convex interval.
It is therefore not necessary to evaluate the entire continuous action space.
It is mathematically sufficient to query the oracle to determine only the minimum safe action $a_{\text{min}}(s)$ and the maximum safe action $a_{\text{max}}(s)$ using a rapid bisection search algorithm.
Because the action space is convex with respect to the safety constraints, any action $a$ that satisfies $a_{\text{min}}(s) \le a \le a_{\text{max}}(s)$ is intrinsically guaranteed to be feasible.

Consequently, it is not necessary to evaluate the entire continuous action space.
It is mathematically sufficient to query the oracle to determine only the minimum safe action $a_{\text{min}}(s)$ and the maximum safe action $a_{\text{max}}(s)$ using a rapid bisection search algorithm.
Because the action space is convex with respect to the safety constraints, any action $a$ that satisfies $a_{\text{min}}(s) \le a \le a_{\text{max}}(s)$ is intrinsically guaranteed to be feasible. 

\section{Safe \Ac{RL} Architecture with Projection Filters}

\subsection{The Safety Filter Mechanism}
During the active \ac{RL} training phase, the agent is strictly prohibited from executing raw actions directly on the hardware or the simulation.
We enforce a deterministic projection filter $P_{s} : \mathcal{A} \rightarrow \mathcal{F}_{\mathcal{A}}(s)$. 

If the agent proposes a raw action $a_t = \pi(s_t)$ that lies within the safe action set $\mathcal{F}_{\mathcal{A}}(s_t)$ derived from the offline \ac{MPC} map, it passes unaltered.
If the action is unsafe, it is projected to the nearest valid point via Euclidean distance, yielding the guaranteed safe action $\overline{a}_t$:
\begin{equation}
    \overline{a}_t = P_{s_t}(a_t) = 
    \begin{cases} 
      a_t, & \text{if } a_t \in \mathcal{F}_{\mathcal{A}}(s_t), \\
      \underset{a' \in \mathcal{F}_{\mathcal{A}}(s_t)}{\arg\min} ||a_t - a'||, & \text{otherwise}.
    \end{cases}
\end{equation}

\subsection{Closed-Loop Evaluation and Reward Shaping}
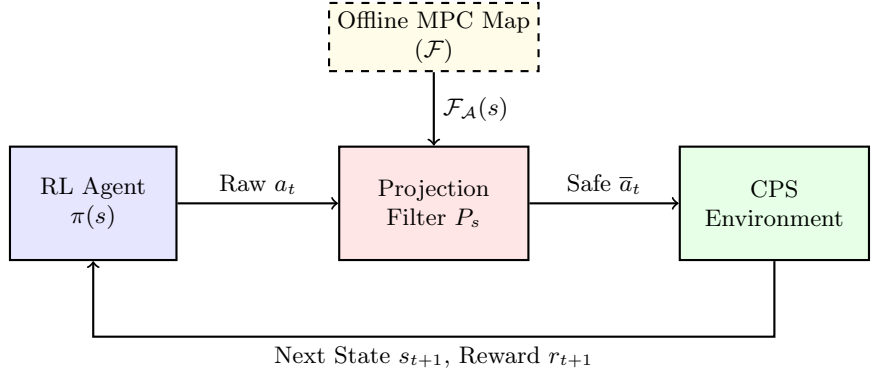
\begin{figure}[htbp]
\centering
\begin{tikzpicture}[node distance=2.5cm, auto, thick]
    % Nodes
    \node [draw, rectangle, fill=blue!10, minimum height=1.5cm, minimum width=2.2cm, align=center] (agent) {RL Agent \\ $\pi(s)$};
    \node [draw, rectangle, fill=red!10, right of=agent, node distance=4.5cm, minimum height=1.5cm, minimum width=2.5cm, align=center] (filter) {Projection \\ Filter $P_s$};
    \node [draw, rectangle, fill=green!10, right of=filter, node distance=4.5cm, minimum height=1.5cm, minimum width=2.5cm, align=center] (env) {CPS \\ Environment};
    \node [draw, dashed, rectangle, fill=yellow!10, above of=filter, node distance=2.2cm, align=center] (oracle) {Offline MPC Map \\ ($\mathcal{F}$)};

    % Edges
    \draw [->] (agent) -- node {Raw $a_t$} (filter);
    \draw [->] (filter) -- node {Safe $\overline{a}_t$} (env);
    \draw [->] (oracle) -- node {$\mathcal{F}_{\mathcal{A}}(s)$} (filter);
    
    % Feedback loop
    \draw [->] (env.south) -- ++(0,-1) -| node[pos=0.25, below] {Next State $s_{t+1}$, Reward $r_{t+1}$} (agent.south);
\end{tikzpicture}
\caption{Architecture of the Safe RL Framework. Unsafe actions proposed by the RL agent are evaluated instantaneously against the current state and projected to the nearest recursively feasible boundary defined by the offline MPC map.}
\label{fig:architecture}
\end{figure}

The interaction between the offline mapped boundaries and the active learning loop is illustrated in \cref{fig:architecture}.
At each time step $t$, the agent evaluates the current state $s_t$ and proposes a raw action $a_t$.
The projection filter intercepts $a_t$ and queries $\mathcal{F}$ to retrieve the instantaneous safety bounds.
Only the guaranteed safe action $\overline{a}_t$ is transmitted to the \ac{CPS} environment.

In our \ac{RL} training setup, closed-loop evaluation is critical to accurately represent the physical control loop.
During the numerical simulation of the environment, the subsequent state $s_{t+1}$ is computed by integrating the system dynamics using the current safe action $\overline{a}_t$.
This newly computed state is then immediately fed back into the policy to sample the next raw action $a_{t+1}$, ensuring a continuous, state-reactive feedback loop.
To encourage instructional learning, the magnitude of the projection $||\pi(s) - \overline{\pi}(s)||$ can be penalized in the \ac{RL} reward function.
This incentivizes the agent to organically learn the system boundaries and output feasible actions natively, reducing reliance on the filter over time.
Furthermore, to prevent hardware-damaging high-frequency oscillation (\enquote{bang-bang} control) common in \ac{DRL}, an action penalty term may be applied to the base reward, computed via the standard deviation of the normalized applied action over a sliding window \cite{SKRHH25}.

\section{Experimental Validation on a Cyber-Physical Testbed}

To validate the theoretical generalized framework, we deployed the algorithm on a representative non-linear laboratory testbed: the Quanser Aero 2. 

\subsection{The Quanser Aero 2 Testbed}
\begin{figure}[htbp]
    \centering
    \includegraphics[width=0.9\linewidth]{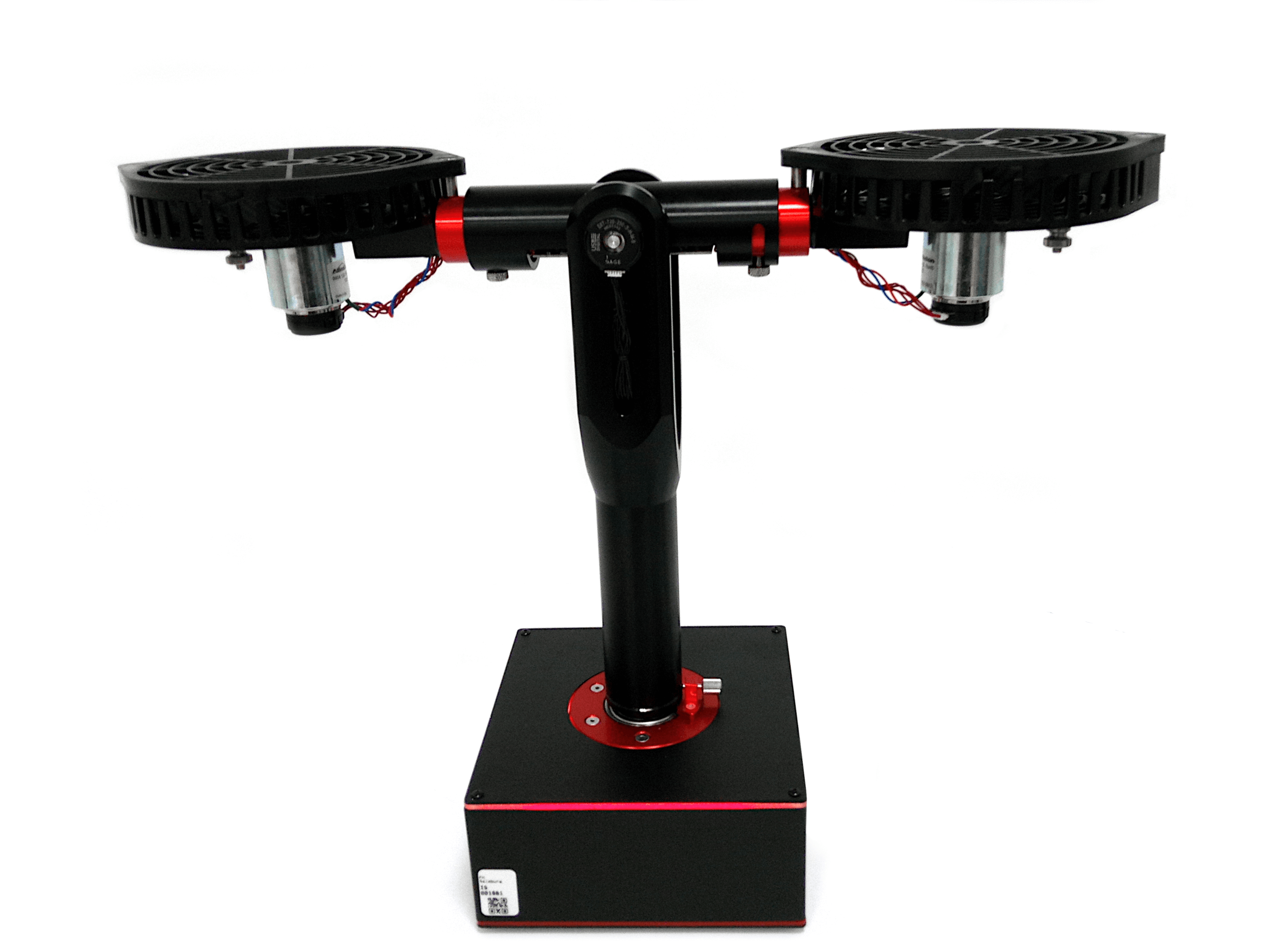}
    \caption{The Quanser Aero 2 experimental setup. The system is configured for 1-DOF pitch control via horizontally mounted motors, with mechanical pitch limits of $\pm60^\circ$.}
    \label{fig:quanser_aero}
\end{figure}

The Aero 2 is a reconfigurable dual-rotor aerospace experiment.
For this validation, we configured the system for \ac{1-DOF} pitch control, mechanically locking the yaw axis (see \cref{fig:quanser_aero}).
Equipped with two motors driving horizontal fans, the system's inputs are simplified to a single voltage $u$, operating the fans in opposition \cite{SRHH24}.

From the principle of angular momentum, the non-linear equations of motion for the pitch angle $\Theta$ and pitch velocity $\omega = \dot{\Theta}$ are derived as:
\begin{align}
    \dot{\Theta} &= \omega, \\
    \dot{\omega} &= \frac{1}{J_p}(-k_d \omega - d_S m g \sin \Theta) + 2 k_u u,
\end{align}
where $d_S$ is the distance from the pivot to the beam's center of gravity, $m$ is the mass, and $g$ is gravity.
The friction coefficient $k_d$, moment of inertia $J_p$, and voltage gain $k_u$ are identified from empirical data \cite{SRHH24}.
The system is subject to strict mechanical pitch constraints of $\pm 60^\circ$ ($\approx \pm 1.05$ rad), defining $\mathcal{X}_{\text{safe}}$.

\subsection{Offline Mapping and System Integration}
\begin{figure}[htbp]
    \centering
    \includegraphics[width=1.0\linewidth]{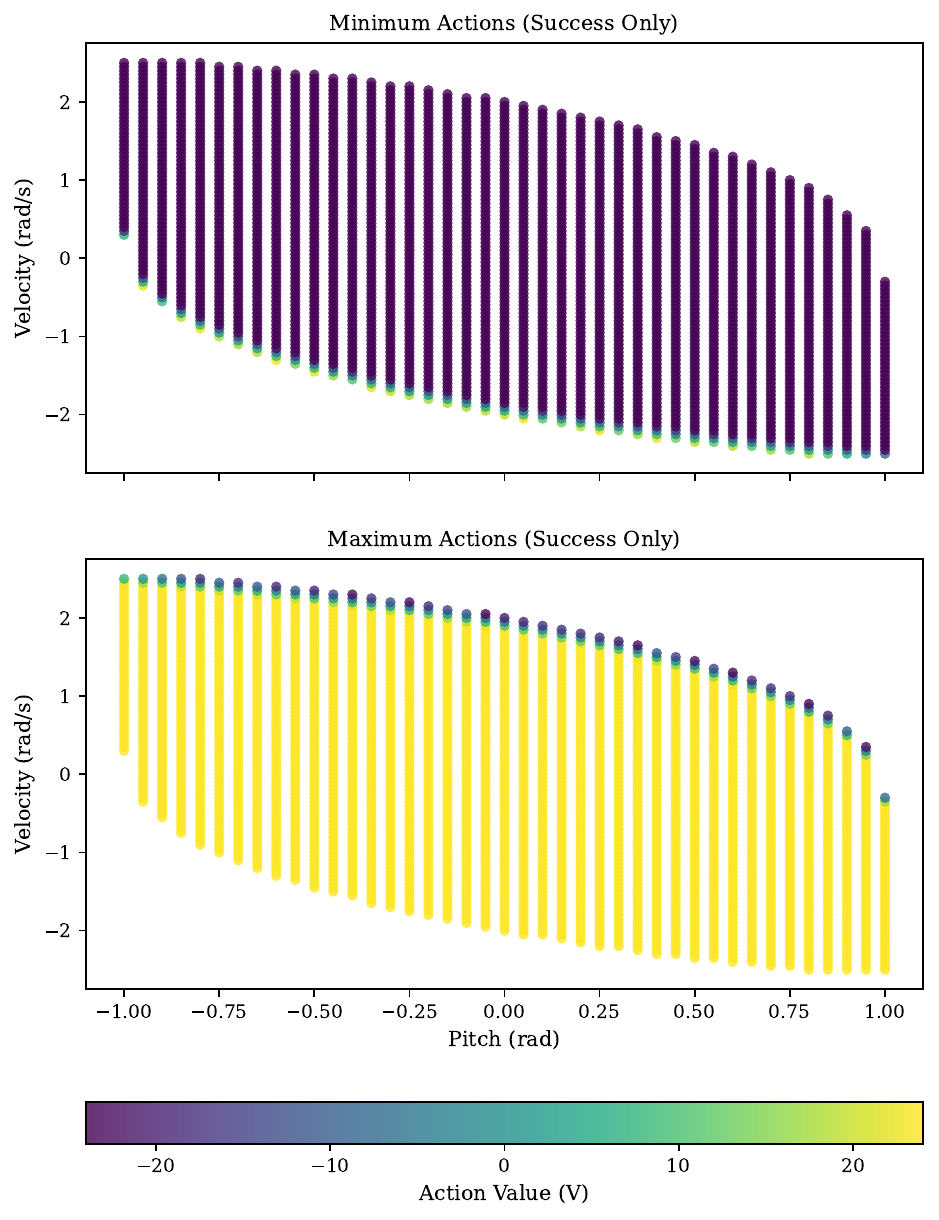}
    \caption{Determined Feasible State Space for the Quanser Aero 2. The top plot shows the minimum required actions for safety, while the bottom plot shows the maximum actions. Due to system convexity, any control input falling between these two boundary surfaces is formally feasible.}
    \label{fig:min_max_actions}
\end{figure}

Using the offline \ac{MPC} approach and our bisection grid search strategy (exploiting the monotonic relationship between rotor voltage and thrust), we sampled the safe operational envelope.
\Cref{fig:min_max_actions} illustrates the resulting feasible state space based on the minimum and maximum action boundaries.
The graphical results vividly demonstrate the abstract concept shown previously in \cref{fig:point_of_no_return}: at high absolute velocities approaching the physical limit, the set of feasible voltages shrinks dramatically to only those capable of maximum reverse thrust.

To bridge the gap between the mathematical synthesis and the Python-based \ac{DRL} training (e.g., Stable Baselines3), we rely on a systematic software architecture detailed in our previous work \cite{SSRHH24}.
The Simulink mathematical model is translated into a compiled \ac{DLL} and wrapped in a Farama Gymnasium interface, enabling high-frequency, zero-shot deployment to the physical hardware via the Quanser Python API.

\subsection{Empirical Results}
Integration of the safety filter allowed the \ac{PPO} \cite{schulman2017proximal} agent to train directly on the simulated and physical testbed.
Empirical observations indicate that the agent successfully explores the extreme boundaries of the state space, achieving high control performance while remaining within the defined feasible bounds for the vast majority of the training process. 

While the theoretical mathematical framework guarantees recursive feasibility and strict constraint satisfaction, empirical deployment on physical hardware revealed rare, intermittent mechanical constraint violations.
These edge-case violations are primarily attributed to unavoidable model inaccuracies, the classic \enquote{Sim-to-Real} gap, where the true physical system dynamics deviate slightly from the nominal mathematical model used by the oracle.
Additional contributing factors include practical implementation artifacts, such as discrete-time sampling errors (where continuous dynamics violate bounds between evaluation steps), minor unmodeled hardware communication latencies, and numerical solver tolerances during the real-time Euclidean projection.

Despite the physical realities of the \enquote{Sim-to-Real} gap, the projection filter operates efficiently at high frequencies.
It successfully prevents the catastrophic, high-velocity exploration failures inherent to unconstrained \ac{DRL}, drastically reducing the violation rate compared to baseline methods and stabilizing the hardware training phase.

\section{Discussion and Future Work}

The proposed methodology robustly combines the safety guarantees of \ac{MPC} with the high-frequency adaptable nature of \ac{RL}.
By mapping the feasible state-action space offline, and executing the policy via an efficient projection filter, we eliminate the real-time computational bottleneck of \ac{MPC} while strictly preventing catastrophic exploration failures.

However, the methodology presents notable limitations.
First, the absolute safety guarantee is strictly dependent on the accuracy of the system dynamics model, i.e., the classic \enquote{Sim-to-Real} gap, as well as the numerical robustness and latency of the real-time hardware execution loop.
Second, while the systematic grid search is effective for the \ac{1-DOF} evaluation, it ultimately suffers from the curse of dimensionality.
For a state-action space of dimension $d$, the number of required grid evaluations scales exponentially, rendering it intractable for highly complex systems.

Future work will focus on addressing these scalability challenges by replacing the grid search with \acp{GP} \cite{rasmussen2003gaussian} for active sampling of the oracle near the isolating boundary $\partial \mathcal{F}$.
Furthermore, we intend to integrate robust \ac{MPC} formulations during the offline phase to systematically account for model discrepancies, and explore \ac{MORL} frameworks to natively balance the trade-offs between aggressive task performance and the distance to the pre-computed safety boundaries.

\section*{Acknowledgment}
Financial support for this study was provided by the Christian Doppler Research Association (CDG) through the Josef Ressel Centre for Intelligent and Secure Industrial Automation, the corresponding WISS Co-project of Land Salzburg, and by the European Interreg project BA0100172 AI4GREEN.

\bibliographystyle{splncs04}
\bibliography{references,jrcisia-published}

\end{document}